\documentclass[runningheads]{llncs}
\usepackage[T1]{fontenc}
\usepackage{needspace}
\usepackage{graphicx}
\usepackage{amssymb}
\usepackage{amsfonts}
\begin{document}
\title{Distilling Knowledge from Large Language Models
into Lightweight Reinforcement Learning Agents
for Autonomous Cyber Operations}
\titlerunning{Distilling LLM Knowledge into Lightweight RL Agents}

\author{
Konur Tholl\inst{1}\orcidID{0009-0008-8078-9764} \and
Fran\c{c}ois Rivest\inst{2}\orcidID{0000-0003-2038-5174} \and
Mariam El Mezouar\inst{2}\orcidID{0000-0002-3317-7051} \and
Adrian Taylor\inst{3}\orcidID{0000-0002-3785-6270} \and
Ranwa Al Mallah\inst{4}\orcidID{0000-0003-3703-9729}
}

\authorrunning{K. Tholl et al.}

\institute{
Department of Electrical and Computer Engineering,\
Royal Military College of Canada, Kingston, Canada
\and
Department of Mathematics and Computer Science,\
Royal Military College of Canada, Kingston, Canada
\and
Defence Research and Development Canada, Ottawa, Canada
\and
Department of Computer and Software Engineering,\
Polytechnique Montreal, Montreal, Canada
}

\maketitle              
\begin{abstract}
Autonomous Cyber Operations (ACO) have become increasingly important for defending enterprise networks as cyber threats continue to evolve in scale and sophistication. ACO applications commonly employ Reinforcement Learning (RL) agents to learn defensive behaviors through direct interaction with environments. However, RL agents typically require extensive exploration during training, often resulting in unstable behavior and poor initial decision-making before converging toward effective defense strategies.

In this work, we investigate the use of a Large Language Model (LLM) to improve autonomous defensive decision-making within an ACO environment. Through prompt engineering rather than environment-specific fine-tuning, we demonstrate that an 8-billion parameter LLM pretrained on cybersecurity data can outperform a baseline RL agent in a modified CybORG CAGE Challenge 2 environment. We then propose an online policy distillation framework that transfers the LLM's defensive policy into a lightweight RL agent containing only 64,910 parameters, reducing model size by several orders of magnitude while maintaining effective defensive capabilities. This provides a potential pathway toward operationalizing the defensive capabilities of computationally expensive frontier cybersecurity models within lightweight, operationally deployable agents.

To evaluate transferability, we additionally construct multiple CybORG scenarios ranging from 4 to 12 hosts and assess the feasibility of the proposed approach across varying network configurations. Furthermore, we systematically evaluate multiple teacher-guided RL stabilization strategies and observe that none consistently surpass the optimized teacher policy, suggesting potential policy-alignment limitations between reward-driven RL optimization and teacher-guided defense strategies.

Our results demonstrate the potential of using cybersecurity-focused LLMs as sources of expertise for autonomous cyber defense, while policy distillation provides a practical path toward operationalizing the defensive capabilities of frontier cybersecurity models within computationally efficient and scalable agents.
\end{abstract}


\Needspace{10\baselineskip}

\section{Introduction}
\label{sect:introduction}
Reinforcement Learning (RL) has emerged as a promising approach for countering continuously evolving adversarial cyber threats. Unlike traditional Machine Learning (ML) techniques, which often require maintaining large labeled datasets that must continually be updated to remain effective against new attack strategies, RL enables agents to learn defensive behaviors through direct interaction with an environment, reducing dependence on manually curated datasets. However, traditional RL approaches present several challenges in the cybersecurity domain. In particular, RL agents often require long training times and must initially perform unfavorable actions in order to learn from their consequences. In cybersecurity environments, where incorrect decisions may result in severe operational impacts such as the compromise of an enterprise network, this exploratory learning strategy can be problematic. 

Previous work has explored integrating a teacher into the RL pipeline, where an external entity with prior knowledge of the field initially guides the agent's actions. As training progresses and predefined criteria are met (e.g., a fixed number of episodes), the teacher's influence is gradually reduced until the agent learns independently through environmental feedback. While this approach can significantly accelerate learning and reduce harmful exploration, it introduces an important limitation: complex tasks like cybersecurity may contain multiple favorable policies. Consequently, the teacher's policy may not align with the policy ultimately reinforced through environmental rewards. Even when both policies are effective, this misalignment may negatively impact training stability and overall agent performance.

One potential solution is to focus directly on learning the teacher's policy, rather than jointly optimizing through the teacher's guidance and environmental feedback. For this approach to be effective, the teacher itself must demonstrate strong defensive decision-making capabilities. Large Language Models (LLMs), particularly those pre-trained on cybersecurity-related data, provide a promising candidate for this teacher role due to their flexible input format, embedded domain knowledge, and reasoning capabilities. However, LLMs typically contain significantly more parameters than lightweight RL agents, resulting in increased computational requirements and slower inference times. Recent advances in autonomous cyber defense increasingly rely on these large frontier cybersecurity models capable of advanced contextual reasoning. However, continuously deploying such models in operational environments remains computationally expensive due to their large parameter counts, increased inference latency, and infrastructure requirements. Consequently, an important open problem is whether the defensive policies of these models can be distilled into compact and computationally efficient agents suitable for scalable and resource-constrained deployments.

In this work, we distill the policy of an LLM pretrained on cybersecurity-related data into a lightweight RL agent and demonstrate that the resulting agent outperforms an RL agent trained solely through interaction with a modified version of the CybORG CAGE Challenge 2 environment \cite{thollmodifycyborg2025}. Rather than relying on environment-specific fine-tuning, we optimize the LLM's defensive behavior through prompt engineering and subsequently transfer this policy into a lightweight RL agent. Furthermore, we evaluate the transferability of the proposed approach across multiple CybORG scenarios ranging from 4 to 12 hosts. We additionally investigate whether teacher-guided RL agents can consistently surpass the optimized teacher policy when transitioning toward independent reinforcement learning. The major contributions for this work are summarized as follows:

\begin{itemize}
    \item \textit{LLM-to-RL policy distillation} We propose a method to efficiently distill the knowledge of an 8 billion parameter cybersecurity-focused LLM into a lightweight 64,910 parameter RL agent while maintaining comparable defensive performance.
    
    \item \textit{Transferability evaluation across network topologies.} We construct multiple simulated network environments and evaluate our proposed approach across varying CybORG scenarios to assess its transferability. 

    \item \textit{Teacher-guided RL stabilization analysis.} We systematically evaluate multiple teacher-guided RL stabilization strategies and demonstrate that none consistently surpass the optimized teacher policy, highlighting potential policy-alignment limitations between reward-driven and teacher-driven defense policies.
\end{itemize}

The rest of this paper is organized as follows: Section \ref{sect:relatedwork} discusses related work in ACO, RL, and teacher-guided learning approaches. Section \ref{sect:methodology} presents the methodology used to implement the proposed solution. Section \ref{sect:evaluation} evaluates the performance and transferability of the proposed approach across multiple CybORG environments. Finally, Section \ref{sect:conclusion} concludes the paper and discusses future research directions.

To support reproducibility, the source code, configurations, prompts, and experimental setup are available in the accompanying GitHub repository at https://github.com/Poly-AIvsAI/LLMDistillationACO.

\section{Related Work}
\label{sect:relatedwork}
In this section, we discuss previous work in ACO, teacher-guided RL, knowledge distillation and LLMs. We then synthesize our findings to motivate the proposed approach and identify the research gap addressed in this work.

\subsection{ACO and RL}
Researchers have increasingly explored RL for applications in ACO, where agents learn defense strategies by directly interacting with their respective environment \cite{sutton_reinforcement_2014}. For RL to be effective in cybersecurity domains, the environment must both realistically model cyber operations and provide meaningful signals that enable the agent to converge towards effective policies. CybORG satisfies these requirements, by providing a cybersecurity environment where blue agents can be trained to defend a simulated enterprise network \cite{baillie_cyborg_2020,kiely2023autonomous}.

Previous work, such as that conducted by Palmer et al., Wiebe et al., and McDonald et al. demonstrate the feasibility of training RL agents within the CybORG environment \cite{palmer_deep_2024,wiebe_learning_2023,mcdonald_competitive_2024}. However, these approaches typically initialize agents without prior domain knowledge, requiring them to learn entirely through environmental exploration. Consequently, agents often exhibit poor initial performance and require substantial training time before converging toward effective defensive policies. In cybersecurity, where incorrect actions may have a detrimental impact on security posture, this exploratory learning process can directly impede ACO's practical adoption.

\subsection{Teacher-Guided RL}
Teacher-guided RL has emerged as a promising approach for mitigating the inherent limitations associated with learning from scratch. One common strategy involves using a teacher model to guide the student's early learning process before gradually reducing the teacher's influence over time. For example, M. Pfeiffer et al., proposed generating a labeled dataset using the teacher to train the student prior to transitioning to conventional RL training \cite{pfeiffer_reinforcedimitationlearning_2018}. More recent work incorporated the teacher's guidance directly within the RL environment through approaches such as reward shaping, feature space modification, action guidance, and auxiliary loss signals rather than through a separated pretraining phase \cite{beikmohammadi_ta-explorerewardshaping_2023,wang_learningactionmasking_2024,tholl_comparative_2025}. 

These approaches demonstrated that teacher guidance can accelerate convergence and improve initial agent performance. However, they still rely on environmental rewards as the primary optimization objective. In complex domains such as cybersecurity, multiple viable defensive policies may exist for achieving favorable outcomes. Consequently, the teacher's policy may not align with the policy ultimately reinforced through environmental feedback. Even when both policies are effective, this mismatch may require the student to partially unlearn the teacher's guidance during later training stages, potentially destabilizing learning and reducing the effectiveness of teacher-guided training. 

\subsection{Knowledge Distillation}
One potential approach for mitigating possible teacher-policy misalignment is to derive the student's training directly from the teacher rather than jointly optimizing through environmental rewards. This concept is not novel in itself and has previously been explored through various forms of knowledge distillation and imitation learning across multiple domains \cite{santos_distilling_2025,pozzi_mitigating_2025,yu_language_2026,agarwal_-policy_2024}. Prior work like those done by Agarwal et al. and Pozzi et. al demonstrated that complex teacher models can effectively transfer knowledge into smaller and computationally efficient student models \cite{agarwal_-policy_2024,pozzi_mitigating_2025}. 

However, these approaches typically do not focus on online interaction with dynamic RL environments. This presents an opportunity to investigate whether policy distillation can be performed during active interaction with a cybersecurity environment while deriving learning signals directly from the teacher policy rather than environmental reward signals. Under this paradigm, the environment continues providing state transitions and observations, while the teacher acts as the primary source of supervisory guidance.

Because the student no longer directly optimizes against environmental rewards, the effectiveness of the overall approach becomes heavily dependent on the quality and consistency of the teacher policy. Consequently, the teacher must demonstrate strong defensive decision-making capabilities, ideally without requiring extensive environment-specific fine-tuning. 

\subsection{LLMs}
LLMs have shown significant promise in cybersecurity due to their ability to recognize complex patterns in text and generate contextually relevant responses \cite{guastalla_applicationllmddos_2024,ali_huntgpt_2023,loevenich_designllmcyborg_2024}. Their large parameter size and extensive pretraining enable them to encode substantial domain knowledge, making them promising candidates for acting as teachers in complex cybersecurity environments. 

However, directly deploying LLMs in operational settings presents several challenges. Their large model sizes require substantial computational resources and typically introduce increased inference times, limitations that are particularly problematic in time-sensitive cybersecurity environments. This motivates investigating whether the capabilities of an LLM can be distilled into a significantly smaller and more efficient RL agent while preserving strong defensive performance.

To the best of our knowledge, no existing LLM has been specifically trained for the CybORG environment. One possible solution would be to follow prior teacher-guided RL approaches where a generalized LLM initially guides training before the agent transitions to learning directly from environmental rewards \cite{tholl_comparative_2025,tholl_large_2026}. However, this may reintroduce the previously discussed policy-alignment challenges. Another option would involve fine-tuning the LLM directly for CybORG; however, this process is computationally expensive and difficult to scale within rapidly evolving cybersecurity environments.

Prompt engineering presents a computationally efficient alternative to fine-tuning, where the underlying model parameters remain fixed while the input prompts are optimized to better align the model with the target environment \cite{matthew_prompt_2024}. Previous work has already demonstrated that prompt engineering can substantially improve LLM performance without modifying model parameters \cite{son_performance_2025}. For example, Son et al. improved Llama3's TruthfulQA performance from 3.2 to 13.65 BLEU using Retrieval-Augmented Generation (RAG), representing a 323\% relative improvement \cite{son_performance_2025}.

\subsection{Discussion}
Prior work demonstrated the feasibility of applying RL to ACO environments such as CybORG; however, these approaches typically require agents to learn entirely through exploration, resulting in poor initial performance and extended convergence times \cite{mcdonald_competitive_2024,palmer_deep_2024,wiebe_learning_2023}. Teacher-guided RL approaches helped mitigate some of these limitations, but continued reliance on environmental reward signals may introduce policy-alignment challenges between the teacher and student \cite{tholl_comparative_2025,tholl_large_2026}. Existing work also demonstrated the potential of leveraging LLMs for cybersecurity-related reasoning and decision-making \cite{guastalla_applicationllmddos_2024,ali_huntgpt_2023,loevenich_designllmcyborg_2024}.

Collectively, these observations motivate investigating whether a large frontier cybersecurity model can be optimized for CybORG through prompt engineering and subsequently distilled into a lightweight RL agent capable of operating more efficiently within simulated enterprise networks. Furthermore, the potential alignment limitations of teacher-guided RL suggest that directly distilling expert-guided defense strategies may provide a more stable alternative to jointly optimizing through environmental reward signals. Finally, previous work in CybORG has largely focused on single simulated environments, creating an opportunity to evaluate transferability across varying network topologies to assess the robustness of the proposed methodology. 

\section{Methodology}
\label{sect:methodology}
This section outlines the phased approach we used in our work. These phases include:
\begin{enumerate}
    \item \textit{Distillation}. We distill the LLM's knowledge into a lightweight RL agent while it actively interacts with the environment. Action masking is employed to ensure the RL agent always performs adequately while a loss signal derived from the teacher's feedback is used to optimize its underlying parameters.

    \item \textit{Prompt engineering}. We use an LLM pretrained on cybersecurity that has already been evaluated in CybORG as the frontier-model teacher \cite{tholl_large_2026}. We then optimize the prompt using a constrained chain-of-thought template with reasoning rules and output constraints.
    
    \item \textit{Transferability}. We create 9 simulated scenarios ranging from 4 to 12 hosts to assess the transferability of our solution.
\end{enumerate}

\subsection{Distillation}
The distillation process we used employed action masking to ensure the RL agent always performed consistently with the teacher \cite{tholl_comparative_2025}. In particular, we manually set the probability of selecting any action not recommended by the teacher to 0:

\begin{equation}
    \pi_{\mathrm{masked}}(a_t)=\frac{M_t(a_t)}{\sum_{a'} M_t(a')}
\end{equation}

where \(M_t(a_t)\) is the masking matrix where every action not recommended by the teacher is 0 and the recommended action is set to 1. We divide by the sum of all elements in the masking matrix \(\sum_{a'} M_t(a')\) to ensure that a valid probability distribution is maintained.

Masking ensures that there is an immediate performance improvement to our RL agent, but it does not update the RL agent's underlying parameters \cite{tholl_comparative_2025}. To distill the LLM's policy into the RL agent itself, we incorporated a loss signal derived from the LLM's recommendation.  In particular, we computed the loss signal as:

\begin{equation}
    L^{teacher}(\theta)=-log(\pi_{\theta}(a_t^{teacher}|s_t))
\end{equation}

where \(L^{teacher}(\theta)\) is the teacher-guided loss function, and \(log(\pi_{\theta}(a_t^{teacher}|s_t))\) is the log probability of selecting the teacher's recommendation in the current policy. If the agent is likely to select the teacher's recommendation in its current policy, the loss is low, whereas if it is unlikely to select the teacher's recommendation, loss will increase exponentially, encouraging the policy to align more closely with that of the teacher's.

We continued this distillation process using the teacher-derived loss signal for 240 episodes before switching to RL-agent only action selection.

\subsection{Prompt Engineering}
We further refine the basic prompt engineering employed in previous work to better align the LLM with the operational constraints and decision-making requirements of the modified CybORG environment \cite{huggingface_cyber-risk-llama,tholl_large_2026,thollmodifycyborg2025}. 

For this, we employed a combination of role grounding, zero-shot learning, rule semantics, and chain-of-thought scaffolding \cite{nashid_retrieval-based_2023,palmer_deep_2024,jawad_deep_2023}. We refer to this approach as chain-of-thought scaffolding instead of the well-known chain-of-thought reasoning because we do not simply tell the LLM to explain its sequence, but instead provide a structured high-level reasoning process intended to guide defensive decision-making. The complete prompt can be found in the accompanying GitHub repository \cite{tholl_distilledgithubforpaper_2026}.

To preserve operational realism and increase generalization, the information included in the prompt was intentionally constrained to information that would realistically be available to human defenders. For example, privileged simulator state and adversarial tactics, techniques and procedures (TTPs) were not included in the prompt.  

\begin{figure}
    \centering
    \includegraphics[width=0.4\columnwidth]{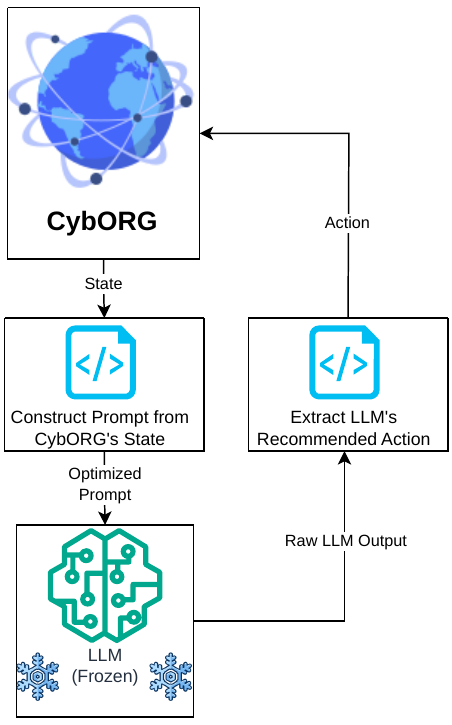}
    \caption[High-Level Prompt Creating.]{Overview of transforming CybORG's state space into a coherent prompt and extracting an executable action from the LLM.}
    \label{fig:creatingPrompt}
\end{figure}

The same procedure as done in Tholl et al. was used to dynamically generate information for the prompt using CybORG's state space as well as extracting the LLM's recommended action using a combination of regex against the existing actions and falling back to semantic similarity if regex failed \cite{tholl_large_2026}. The high-level process for creating the prompt and extracting an action from the LLM is shown in Fig. \ref{fig:creatingPrompt}.

\subsection{Transferability}
To assess the transferability of the proposed approach across different environments, we created nine additional scenarios ranging in complexity from 4 to 12 hosts. A similar B-line agent was used on the red side to infiltrate the networks, with minor modifications to make it function for the various simulated networks \cite{baillie_cyborg_2020}. 

To keep the assessment fair, the same 8 billion parameter Cyber Risk Llama LLM was used with identical prompt structures \cite{huggingface_cyber-risk-llama}. The only hyperparameter that changed throughout was the time at which we stopped distilling the LLM's knowledge into the RL agent, as it was found that less time for knowledge transfer was required for smaller network topologies.

\Needspace{10\baselineskip}
\section{Evaluation}
\label{sect:evaluation}
In this section, we present, evaluate, and interpret the results of our work with respect to their implications for autonomous cyber defense and RL-guided defensive learning. In particular, this section covers:

\begin{itemize}
    \item \textit{Prompt engineering.} Evaluating our optimized prompt against the baseline used in \cite{tholl_large_2026}.

    \item \textit{Distillation.} Assessing the feasibility of distilling the teacher's performance into an RL agent and comparing it against a baseline RL agent and the teacher-guided approach used in \cite{tholl_large_2026}.
    
    \item \textit{Learning stabilization.} Investigating whether teacher-guided RL strategies can consistently surpass the optimized teacher policy during independent reward-driven learning.
    
    \item \textit{Transferability.} We evaluate our solution's performance across multiple simulated networks of varying complexity ranging from 4 to 12 hosts.
\end{itemize}

\subsection{Prompt Engineering}
\begin{figure}
    \centering
    \includegraphics[width=0.8\columnwidth]{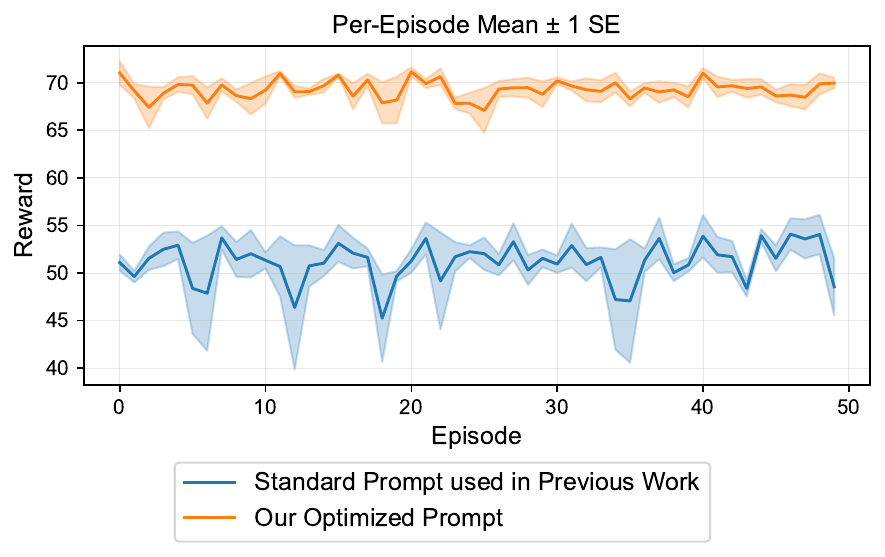}
    \caption[Comparing Prompts]{Evaluation of the Standard and Optimized Prompt across 10 independent runs for 50 episodes. Per-episode mean reward with a ±1 standard error.}
    \label{fig:comparingPrompts}
\end{figure}

As discussed in Section \ref{sect:methodology}, we further refined the prompt from what was used in Tholl et. al's work \cite{tholl_large_2026}. This was an iterative process of incrementally modifying the chain-of-thought scaffolding to guide the LLM's decision-making until superior performance was observed. We present the performance of the optimized prompt versus the standard one in Figure \ref{fig:comparingPrompts}.

The metric used to evaluate the performance of the prompts is the reward obtained by the LLM with respect to CybORG's reward signals. We can see that using the optimized prompt yields an average reward of roughly 70, corresponding to an approximate performance increase of 35\% with a notably lower standard error compared to the baseline prompt used previously. The constrained chain-of-thought scaffolding likely improved policy consistency by reducing ambiguous action selection and explicitly structuring the LLM's defensive prioritization process around operationally relevant features such as host criticality and suspicious activity.

\subsection{Distillation}
After establishing an optimal prompt, we then proceeded to distill the 8 billion parameter LLM's policy into the lightweight RL agent using the methodology discussed in Section \ref{sect:methodology}. We compare our results against a baseline Proximal Policy Optimization (PPO) agent, the same on-policy RL algorithm that has repeatedly shown success in previous ACO work \cite{wiebe_learning_2023,mcdonald_competitive_2024}. We also compare the distilled agent's performance against a teacher-guided agent using the same teacher-derived loss signal to guide initial training. The difference is that rather than solely learning from the teacher, we decrease the influence of the teacher's auxiliary loss signal and gradually transition to learning solely from the environment's feedback.

\begin{figure}
    \centering
    \includegraphics[width=0.8\columnwidth]{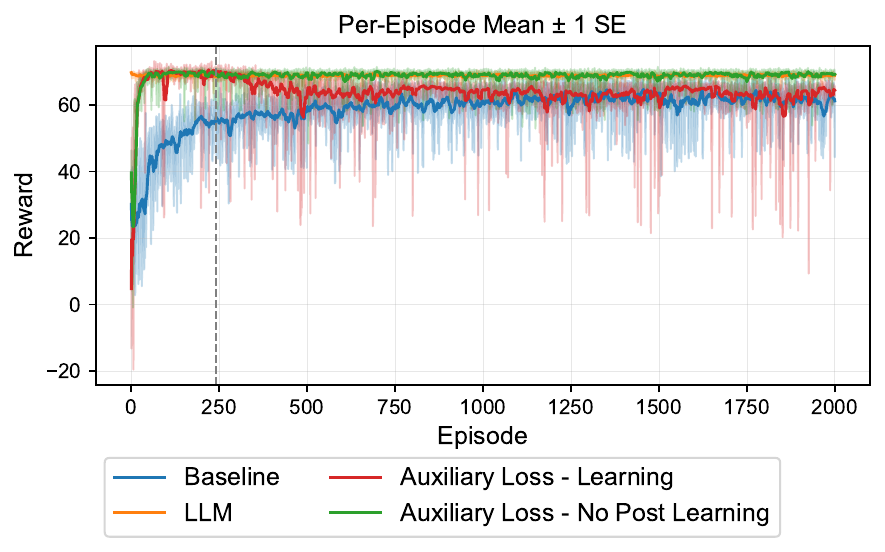}
    \caption[Distilling Knowledge]{Comparing distillation with teacher-guided RL. Per-episode mean reward with a ±1 standard error for 2,000 episodes.}
    \label{fig:distillingKnowledge}
\end{figure}

We can see in Figure \ref{fig:distillingKnowledge} that the distilled agent (without post-distillation learning) outperforms both the baseline and teacher-guided agent. We completely remove the teacher's guidance for our distilled agent by episode 240, while gradually decreasing it for the teacher-guided agent. While the teacher-guided agent has comparable initial performance, it deteriorates noticeably and eventually converges with the baseline when learning from the environmental feedback, resulting in a policy misalignment between the LLM policy and the policy learned solely from CybORG's signals.

\begin{figure}
    \centering
    \includegraphics[width=0.8\columnwidth]{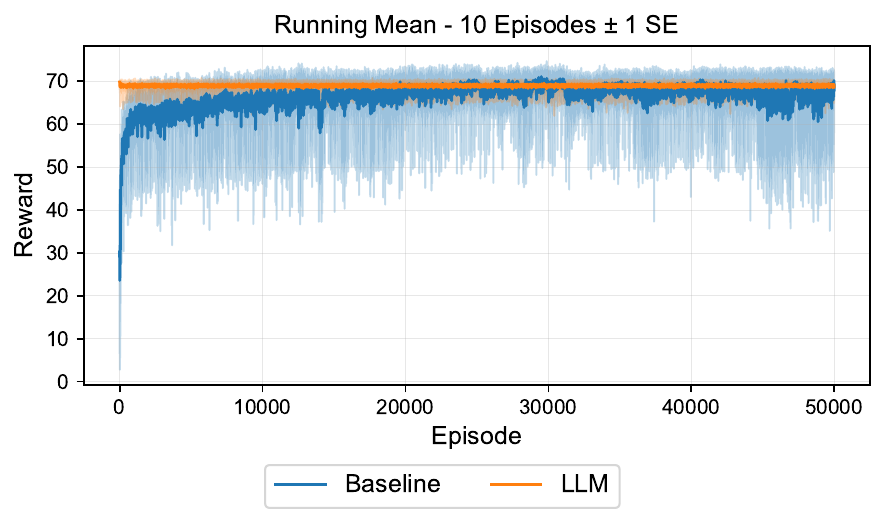}
    \caption[Comparing to Baseline]{Comparing the distilled agent against the baseline PPO agent across 10 independent runs using a 10-episode running average with a ±1 standard error for 50,000 episodes.}
    \label{fig:baselineComparison}
\end{figure}

Our next question was whether the baseline PPO agent would ever surpass the performance of our distilled agent. For this, we ran our baseline for 50,000 episodes (1.6 million timesteps) as shown in Fig. \ref{fig:baselineComparison}. 

From the Figure, we can see that while individual runs surpass our distilled agent's performance, the mean performance never appears to stabilize beyond our agent. There are points after 23,000 episodes where the mean performance of the baseline temporarily surpasses our distilled agent, but the performance does not remain stable throughout training, whereas our distilled agent remains consistently stable throughout the 50,000 episodes.  

\subsection{Transferability}
To assess the transferability of our solution, we created 9 additional CybORG environments ranging from 4-12 hosts, ensuring that we modified the red agent for each respective environment to optimize its attack trajectory.

For this implementation, we also incorporated action masking during the distillation phase to ensure the agent's performance remained consistent with the teacher policy from the beginning. The LLM's prompt and every hyperparameter were kept constant with the exception of the transition point between distillation and independent RL, with environments containing more hosts requiring a longer distillation phase. The results of our transferability evaluation can be found in Fig. \ref{fig:evaluatingTransferability} under Appendix \ref{app:transferability}.

From Fig. \ref{fig:evaluatingTransferability}, we can see that our methodology for distilling an LLM's policy into a lightweight RL agent transfers reasonably well across the evaluated CybORG environments. The 7 host environment represents an exception, where the baseline RL agent starts to outperform the distilled agent by roughly episode 700 with respect to mean performance. The 5 host, 6 host, 8 host, and 9 host scenarios all show that the baseline performs similarly to the LLM by roughly episode 1250; with the baseline showing similar performance by episode 500 for the 5 host environment with respect to mean-performance. It should be noted that the standard error for the distilled agent is much smaller than the baseline for each of the scenarios, demonstrating lower variability across runs, a property that is desirable in operational cybersecurity environments where stable defensive behavior is important.

\subsection{Improving Post-Learning Performance}
As shown in Fig. \ref{fig:baselineComparison}, when we gradually transition to environmental learning, the performance degrades from the baseline LLM. As such, we attempted various techniques to rectify this drop in performance from teacher-guided to independent RL. Specifically, seven independent techniques were employed:
\begin{itemize}
    \item \textit{Adding the LLM's recommendation to the critic loss.} PPO is an on-policy RL algorithm which contains a critic network that computes a scalar reward signal for helping guide the policy \cite{schulman_proximal_2017}. Here, we incorporate the LLM's guidance not only as an auxiliary loss signal for the actor network, which computes the policy directly, but for the critic network as well.
    Because the learning dynamic is entirely different for the critic network, we opted to integrate the teacher's impact by making it train more strongly on states that were produced from teacher-recommended actions. In particular, we used the following:
    
    \begin{equation}
        L_{LLM}(\phi)=\mathbb{E}[\frac{1}{n}\sum^{n}_{i=1}M_t(a_{i})*(ret_{i}-V{\phi}(s_{i}))^{2}]
    \end{equation}
    
    where \(n\) is the number of samples in the batch, \(M_t(a_i)\) is the masking matrix applied, setting the action to 1 if it came from the LLM recommendation, and 0 otherwise. The last two terms are standard in PPO critic learning: \(ret_i\) are the standard returns computed using the Generalized Advantage Estimate (GAE) and \(V{\phi}(s_{i})\) being the critic network's output for the current state \cite{schulman_proximal_2017}.

    We then combined this auxiliary loss term with the critic's original loss to incorporate the LLM into the critic network's learning process.

    \begin{equation}
        L(\phi)=\sigma L^{env}(\phi)+(1-\sigma)L^{teacher}(\phi)
    \end{equation}

    where \(\sigma\) is gradually increased, reducing the teacher's impact as the agent transitions to independent learning. 
    
    \item \textit{Initializing the critic network with a pretrained one}. To validate whether the problem was the critic potentially lagging behind the actor network, we attempted to initialize the critic network with one that was trained for 10,000 episodes using the baseline PPO agent.
    
    \item \textit{Dynamically adjusting the actor and critic learning rates (LRs).} We dynamically adjusted the critic and actor learning rates to facilitate a smoother transition from teacher-guided to independent RL. In particular, we increased the critic learning rate from \(1.6e^{-3}\) to \(3.2e^{-3}\) to estimate the returns of the LLM-recommended actions, then gradually decreased it to \(0.8e^{-3}\) during the transition to independent RL. The actor network's learning rate was also decayed from \(1.6e^{-3}\) to \(0.8e^{-3}\) during the transition. 
    
    \item \textit{Adding extra critic epochs.} We attempted to double the epochs for the critic network during the transition from teacher-guided to independent RL to enable the PPO agent to quickly evaluate any new, unseen states.
    
    \item \textit{Incorporating the LLM's guidance as a distribution.} Instead of computing the loss signal as the probability of selecting the LLM's single recommendation in the RL agent's policy, we mapped the LLM's raw logits into a distribution across possible actions and used KL divergence to compute the loss \cite{cui_kl_2025}.
    
    \item \textit{Stopping critic learning.} We ceased the learning of the critic network only during the transition to independent RL, while keeping the actor network.
    
    \item \textit{Decay by a multiplicative factor.} Instead of linearly decreasing the impact the teacher has on the agent's training, we decreased it by a multiplicative factor. In particular, we used a multiplicative decay factor of 0.99 per training interval instead of a linear subtraction. 
\end{itemize}

\begin{figure}
    \centering
    \includegraphics[width=0.8\columnwidth]{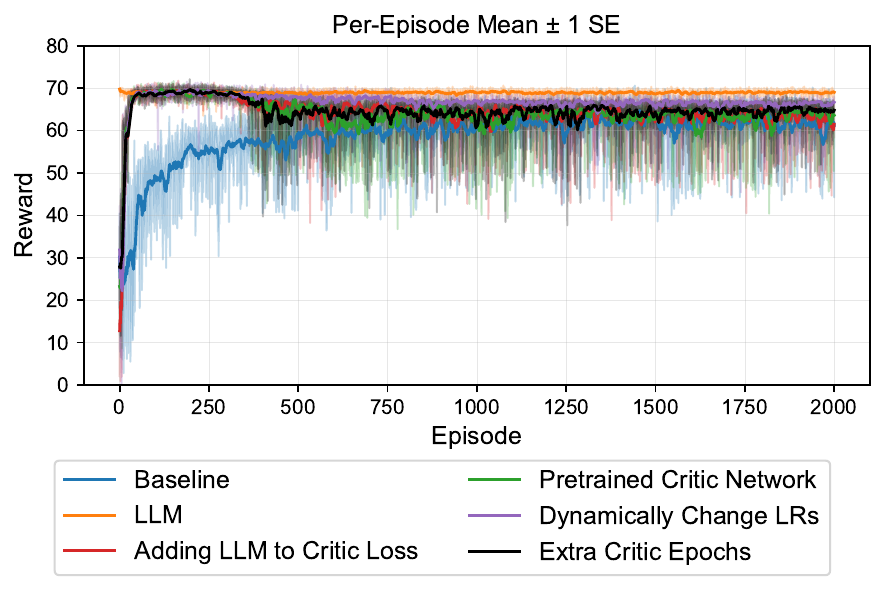}
    \caption[Deficiency Slns]{First four attempts to increase performance beyond the LLM with the optimized prompt. Techniques include: adding the LLM's feedback to the critic's loss, initializing the critic as a pretrained model over 10,000 episodes, dynamically changing the actor and critic LRs, and adding extra learning epochs to the critic during the transition from teacher-guided to independent RL. Plots show a mean reward after applying a 10-episode running average with a ±1 standard error over 2,000 episodes.}
    \label{fig:stabSlnAttempts1}
\end{figure}

\begin{figure}
    \centering
    \includegraphics[width=0.8\columnwidth]{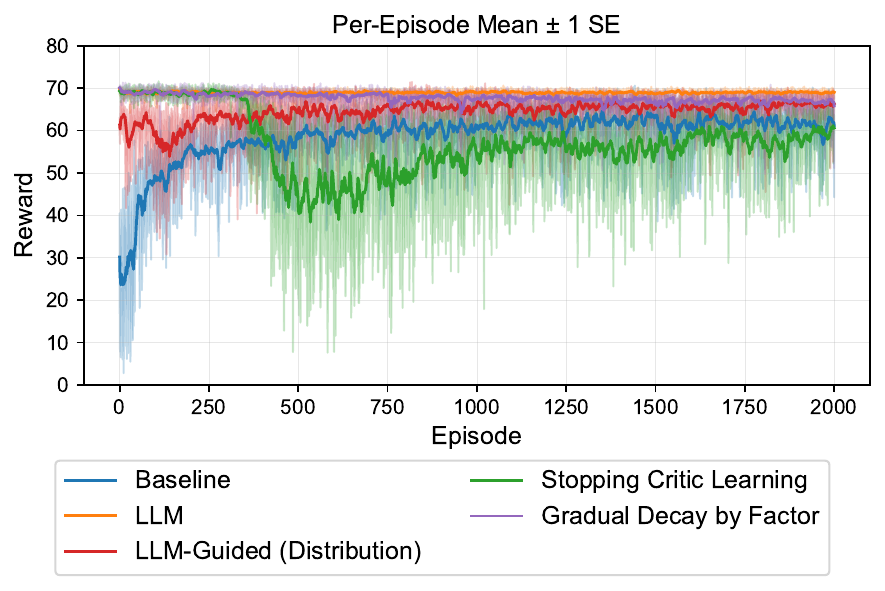}
    \caption[Deficiency Slns 2]{Next three attempts to increase performance beyond the LLM with the optimized prompt. Techniques include: incorporating the LLM's guidance as a distribution (instead of a single action), stopping the critic learning after the transition to independent RL, and decaying the teacher's impact by a multiplicative factor instead of a linear constant. Plots show a mean reward after applying a 10-episode running average with a ±1 standard error over 2,000 episodes.}
    \label{fig:stabSlnAttempts2}
\end{figure}

We present the results of the seven techniques discussed above in Figs. \ref{fig:stabSlnAttempts1} and \ref{fig:stabSlnAttempts2}.

From Fig. \ref{fig:stabSlnAttempts1}, we can see that adding the LLM to the critic loss, initializing the critic as a pretrained network, and adding extra training epochs to the critic network exhibit the same behavior, quickly converging to the teacher's performance and then declining at roughly the same rate until they converge with the baseline by roughly episode 2,000. Dynamically changing the learning rates for the critic and actor network slow down the decline from teacher performance, simply because we're decreasing the extent to which the policy can deviate from the teacher during the transition to independent RL.

In Fig. \ref{fig:stabSlnAttempts2}, stopping the critic learning after the transition to independent RL yields a very noticeable drop in performance from episodes 350 to 500, likely due to being unable to properly quantify any state not directly observed during the teacher-guided phase. Gradually decaying the LLM's influence by a multiplicative factor instead of a linear constant shows similar performance to modifying the learning rate; the performance decreases during the transition to independent RL, just at a slower rate. Finally, computing the LLM's auxiliary loss signal by mapping its raw logits into a distribution over the action space yields poorer initial performance as the policy is now sampled from a distribution rather than directly from the LLM's highest-confidence recommendation; however, this approach shows increased performance after the transition to independent RL, but fails to converge to the LLM's baseline performance.

Overall, none of the seven solutions discussed above to stabilize the teacher-guided RL techniques discussed in the work by Tholl et. al consistently surpassed the LLM's baseline performance \cite{tholl_large_2026}.  

\subsection{Discussion}
We have shown that by shifting focus to prompt engineering, a pretrained and generalized cybersecurity-focused LLM can outperform RL agents specifically trained within the CybORG environment. Furthermore, we demonstrated that the LLM's superior defensive policy can be distilled into a lightweight RL agent in only 240 episodes. These findings suggest that policy distillation may provide a practical pathway for compressing the defensive policies of large cybersecurity-focused frontier models into lightweight agents capable of operating under significantly reduced computational constraints. This is particularly valuable in operational cybersecurity settings, where continuous deployment of large models can be impractical due to increased latency, infrastructure demands, or resource limitations.

We also investigated whether an RL agent guided by the LLM could eventually surpass the teacher policy through independent reward-driven learning by implementing seven modifications to existing teacher-guided RL techniques discussed in previous work; however, none of these approaches consistently surpassed the optimized teacher policy \cite{tholl_comparative_2025,tholl_large_2026,wang_boostinginstruccomprehension_2025,wang_learningactionmasking_2024,beikmohammadi_ta-explorerewardshaping_2023}. Collectively, these findings suggest that environmental reward optimization within CybORG may encourage policies that diverge from expert-guided defensive behavior, even when the resulting policies achieve competitive reward performance.

\Needspace{10\baselineskip}
\section{Conclusion}
\label{sect:conclusion}
Reinforcement Learning (RL) has demonstrated significant promise for Autonomous Cyber Operations (ACO); however, conventional RL approaches remain limited by their dependence on environmental reward signals and their requirement to learn effective defense strategies through exploration from initially untrained policies \cite{wiebe_learning_2023,mcdonald_competitive_2024,baillie_cyborg_2020}. While teacher-guided learning has mitigated some of these concerns, they remain susceptible to policy misalignment between the teacher and the policies reinforced through environmental feedback. 

In this work, we investigated an alternative paradigm, focused on directly optimizing the teacher's policy, and distilling that policy into a lightweight RL agent. Specifically, we demonstrated that a cybersecurity-focused Large Language Model (LLM), optimized through prompt engineering rather than environment-specific fine-tuning, can outperform baseline RL agents within CybORG. We further showed that this performance can be distilled into a compact RL agent several orders of magnitude smaller than the LLM with no noticeable drop in performance across multiple simulated environments. 

\subsection{Contributions}
This study has made several contributions to the fields of autonomous cyber defense, RL, and LLM-guided cybersecurity:

\begin{itemize}
    \item \textbf{Online LLM-to-RL policy distillation.} We proposed an online-distillation framework capable of transferring the policy of an 8-billion parameter LLM into a lightweight RL agent containing 64,910 parameters (approximately 0.0008\% of the teacher model).

    \item \textbf{Transferability evaluation across network topologies.} We evaluated the distilled policy across multiple CybORG scenarios ranging from 4 to 12 hosts, demonstrating that the proposed methodology transfers  reasonably across the evaluated CybORG network topologies.

    \item \textbf{Systematic evaluation of teacher-guided stabilization techniques.} We implemented and evaluated multiple teacher-guided RL stabilization strategies and showed that none consistently surpass the optimized teacher policy, highlighting potential policy-alignment limitations between the environment's reward signals and the teacher's defense behavior.
\end{itemize}

\subsection{Limitations}
Although this work has made meaningful contributions to autonomous cyber defense, several limitations should be considered:

\begin{itemize}

\item \textbf{Environment realism.} While CybORG provides a valuable benchmark for ACO research, it remains a simulated abstraction of operational enterprise environments. Simplified attacker behavior, constrained action and observation spaces, reduced environmental noise, and limited operational complexity restrict the direct applicability of learned defense policies to real cybersecurity settings.  

\item \textbf{Reward-based evaluation metrics.} The primary metric used to evaluate the success of this work was CybORG's scalar reward signal. Although useful for standardized benchmarking, these rewards abstract the agent's underlying behavior, and may not fully capture optimal defensive behavior.

\item \textbf{Dependence on teacher.} The proposed framework distills behavior from a static LLM policy, rather than discovering novel defensive strategies. Consequently, any deficiencies inherent to the teacher, including hallucinations and sub-optimal decision-making, may propagate to the distilled agent. 

\end{itemize}
\subsection{Future Work}
The limitations identified in this work motivate several promising research directions:

\begin{itemize}
    \item \textbf{Reward signal misalignment} Future work should investigate whether current cybersecurity reward structures adequately capture desirable defense behavior and explore alternative evaluation paradigms to quantify the effectiveness of policies beyond scalar rewards.  
    
    \item \textbf{Multi-teacher distillation.} The current framework distills knowledge from a single LLM policy. Incorporating multiple teacher models may improve policy robustness and enhance generalization across different cybersecurity environments.
   
    \item \textbf{Reasoning-aware distillation.} The work distills action-selection behavior. Future approaches could incorporate intermediate reasoning representations or knowledge-graph reasoning to improve policy transfer fidelity.
   
    \item \textbf{Adversarial robustness}. While this work focused on autonomous cyber defense, it did not investigate attacks against either the teacher or student models themselves. Future work should evaluate policy robustness against adversarial techniques such as observation poisoning and prompt injection. 
   
    \item \textbf{Operational Realism.} Future environments should incorporate operational realism through partial observability, increased benign activity, richer host artifacts, and more adaptive attacker behavior.

\end{itemize}

Overall, this work suggests that teacher-guided policy distillation may offer a viable pathway toward operationalizing the defensive capabilities of large frontier cybersecurity models through lightweight, resource-efficient, and scalable autonomous cyber-defense agents.

\bibliographystyle{splncs04}
\bibliography{skeleton}

\appendix

\Needspace{20\baselineskip}
\section{Transferability}
\label{app:transferability}
\begin{figure}[!b]
    \centering
    \includegraphics[width=\columnwidth]{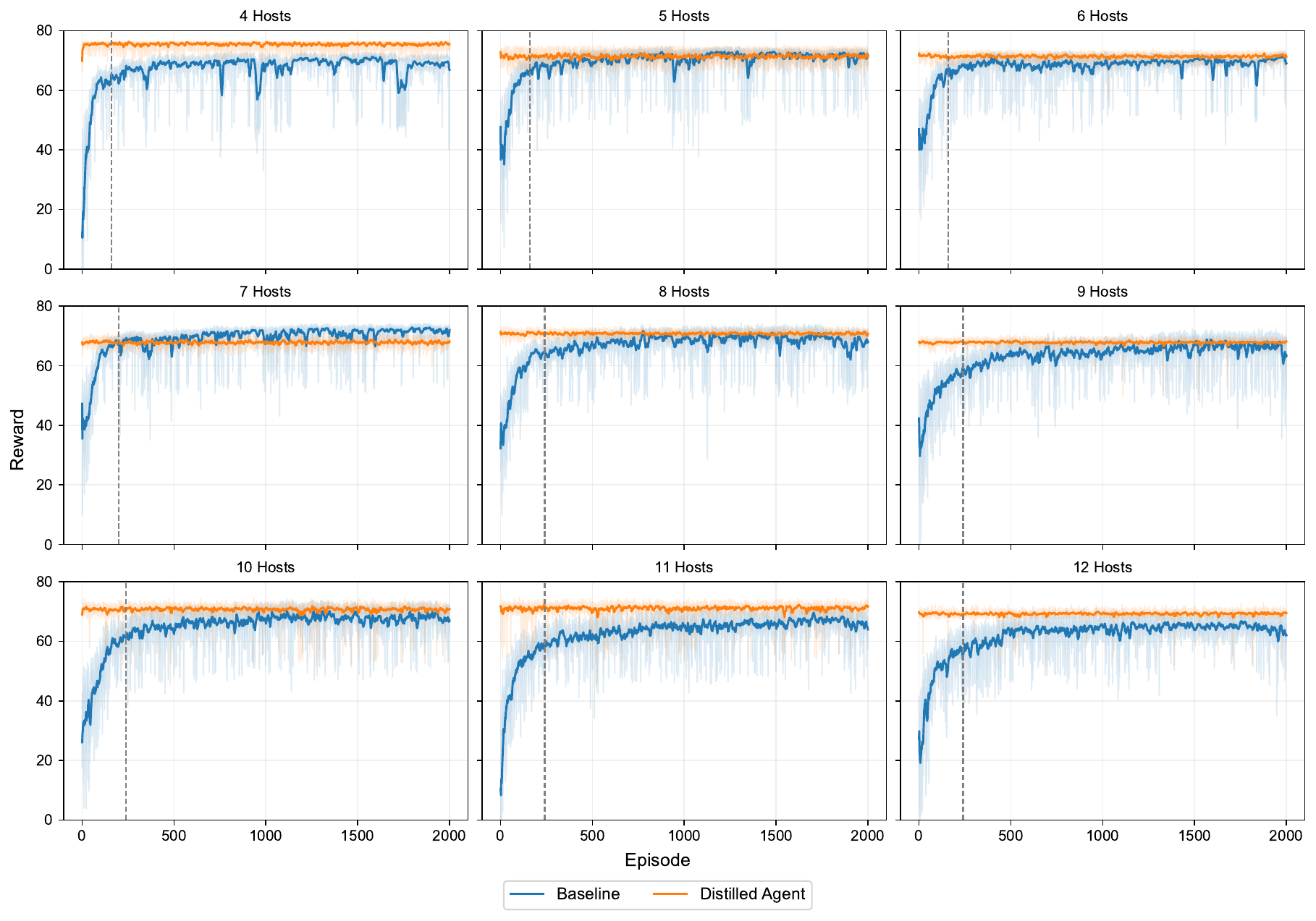}
    \caption[Transferability Evaluation]{Evaluation of the LLM-distilled agent against the baseline agent across different CybORG scenarios ranging from 4 to 12 hosts. The dotted line denotes the point at which the distilled agent has transitioned to acting independently of the LLM. The results are across 10 independent runs, with a 10-episode running average and a ± 1 standard error.}
    \label{fig:evaluatingTransferability}
\end{figure}

As discussed in Sections \ref{sect:methodology} and \ref{sect:evaluation}, we evaluated the transferability of the distilled agent by creating various scenarios in CybORG ranging from 4 to 12 hosts \cite{baillie_cyborg_2020}. The results of these experiments can be found in Figure \ref{fig:evaluatingTransferability}.

\end{document}